\documentclass[twoside,magazine]{IEEEtran}
\usepackage{makecell}
\usepackage{hyperref}
\usepackage{array}
\usepackage{graphicx,amssymb,amsmath}
\usepackage{multicol}
\usepackage[noadjust]{cite}
\usepackage{setspace}
\usepackage{subfig}
\usepackage{graphicx}
\usepackage{float}
\usepackage {url}
\usepackage{stfloats}
\usepackage{amsthm,pifont}
\usepackage{flushend}
\usepackage{cases,subeqnarray}
\usepackage{bm,multirow,bigstrut}
\usepackage{amsmath, amsthm, amssymb}
\usepackage{textcomp}
\usepackage{latexsym,bm}
\usepackage{booktabs}
\usepackage{xcolor}
\usepackage{mathtools}
\usepackage{dsfont}
\usepackage{extarrows}
\usepackage{epsfig}
\usepackage{epsfig}
\usepackage{epstopdf}
\usepackage[noend]{algpseudocode}
\usepackage{algorithmicx,algorithm}
\usepackage{xspace}
\usepackage{listings}
\usepackage{xcolor}
\definecolor{commentcolor}{RGB}{85,139,78}
\definecolor{stringcolor}{RGB}{206,145,108}
\definecolor{keywordcolor}{RGB}{0,0,128}
\definecolor{backcolor}{RGB}{220,220,220}
\theoremstyle{plain}

\theoremstyle{plain}

\usepackage{amsmath}

\IEEEoverridecommandlockouts

\begin{document}
\title{Guiding AI-Generated Digital Content with Wireless Perception
}
\author{Jiacheng Wang, Hongyang Du, Dusit~Niyato,~\IEEEmembership{Fellow,~IEEE}, Zehui Xiong, Jiawen Kang, \\ Shiwen Mao~\IEEEmembership{Fellow,~IEEE}, and Xuemin~(Sherman)~Shen,~\IEEEmembership{Fellow,~IEEE}
 \thanks{J.~Wang, H.~Du and D. Niyato are with the School of Computer Science and Engineering, Nanyang Technological University, Singapore (e-mail: jiacheng.wang@ntu.edu.sg, hongyang001@e.ntu.edu.sg,  dniyato@ntu.edu.sg).}

\thanks{Z. Xiong is with the Pillar of Information Systems Technology and Design, Singapore University of Technology and Design, Singapore (e-mail: zehui\_xiong@sutd.edu.sg)}
 \thanks{J. Kang is with the School of Automation, Guangdong University of Technology, China. (e-mail: kavinkang@gdut.edu.cn)}
 \thanks{S. Mao is with the Department of Electrical and Computer Engineering, Auburn University, Auburn, USA(e-mail: smao@ieee.org)}
    \thanks{X. Shen is with the Department of Electrical and Computer Engineering, University of Waterloo, Canada (e-mail: sshen@uwaterloo.ca).}

}

\maketitle
\vspace{-1cm}
\begin{abstract}
Recent advances in artificial intelligence (AI), coupled with a surge in training data, have led to the widespread use of AI for digital content generation, with ChatGPT serving as an representative example. Despite the increased efficiency and diversity, the inherent instability of AI models poses a persistent challenge in guiding these models to produce the desired content for users. In this paper, we introduce an integration of wireless perception (WP) with AI-generated content (AIGC) and propose a unified WP-AIGC framework to improve the quality of digital content production. The framework employs a novel multi-scale perception technology to read user’s posture, which is difficult to describe accurately in words, and transmits it to the AIGC model as skeleton images. Based on these images and user’s service requirements, the AIGC model generates corresponding digital content. Since the production process imposes the user’s posture as a constraint on the AIGC model, it makes the generated content more aligned with the user’s requirements. Additionally, WP-AIGC can also accept user’s feedback, allowing adjustment of computing resources at edge server to improve service quality. Experiments results verify the effectiveness of the WP-AIGC framework, highlighting its potential as a novel approach for guiding AI models in the accurate generation of digital content.
\end{abstract}

\IEEEpeerreviewmaketitle
\section{Introduction}
The spectacular growth of various types of data, upgradation of hardware, and the advancement of artificial intelligence (AI) algorithms, has led to the emergence of AI generated content (AIGC), which can imitate human behavior to create digital content~\cite{li2018ictu}. Specifically, AIGC refers to the artificial intelligence enabled methods (able to automatically produce, manipulate, and modify multi-modal digital content) and the corresponding generated content~\cite{du2023enabling}. Due to the ability of automatically producing various kinds of high-quality digital content, the AIGC is gaining increasing attention, especially with the rapid integration of the physical world and virtual digital world. 

At the function and application levels, AIGC  enables autonomous content creation through AI~\cite{lugmayr2022repaint}, and boosts the development of various applications. Taking the virtual interactive game in Metaverse as an example, AIGC can generate avatars and create the corresponding scenarios according to users’ requirements, thereby constructing a complete virtual world for users to explore. During this process, the user's needs, e.g., prompts, can be transmitted to the AIGC model through various ways such as voice and text~\cite{thorp2023chatgpt}. Yet, some information is challenging to convey through words, such as the user's posture in the physical world. One feasible solution is to use cameras, such as Kinect, to capture the user's image, which is then combined with the user's requirements and fed into the AIGC model to generate digital content. Nevertheless, prolonged use of the camera may raise privacy concerns, particularly with the rise of face scan payment. Meanwhile, the image transmission can consume significant transmission resources, such as bandwidth. 

\begin{figure*}[t]
	\centering
	\includegraphics[width=0.96\textwidth]{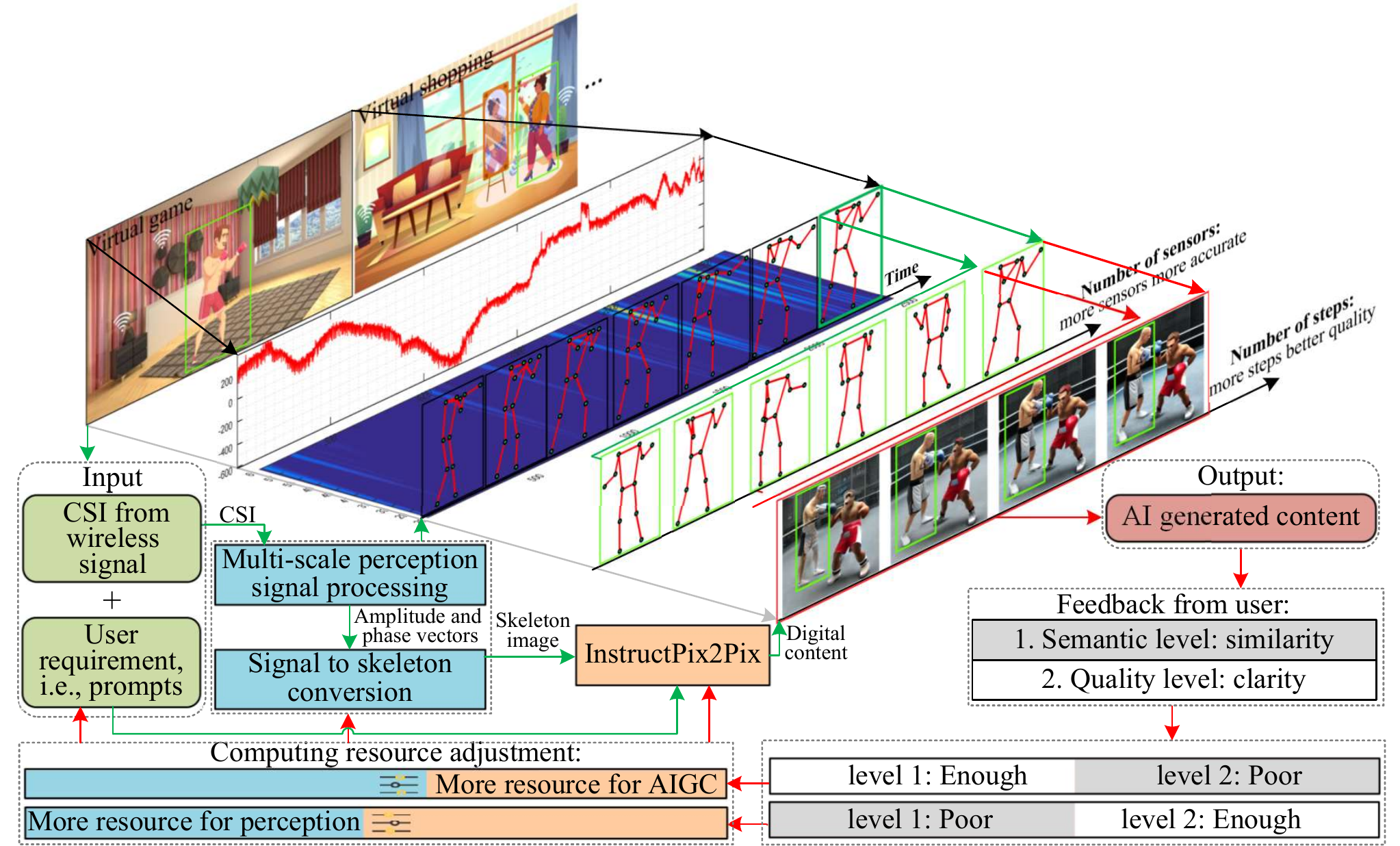}%
	\caption{The overall structure of WP-AIGC. During the operation process, the CSI extracted from wireless signal is first processed, and then a pre-trained network is used to transform the processed CSI into a skeleton image. On this basis, the skeleton image, along with the user's requirements (e.g., prompts), is passed to the AIGC model to generate digital content, which is finally presented to the user in the form of images or other media. Users assess the generated digital content based on their experience, such as whether the virtual character's posture matches their actual posture, and provide feedback to WP-AIGC. Based on feedback received, WP-AIGC adjusts the computing resources at edge server to further enhance the quality of services (QoS).}
	\label{F1}
\end{figure*}

Wireless perception is a suitable alternative to cameras for providing AIGC perception support~\cite{guo2019signal}. The conventional wireless perception demands the deployment of numerous sensor nodes. However, the popularity of smart wireless devices, penetration of wireless networks, and the advances in integrated sensing and communications (ISAC) technologies make the ubiquitous user perception possible. In particular, when the idea of crowdsensing is brought in, more sensors are combined to complete a task, which can further boost the perception quality~\cite{lashkari2018crowdsourcing}. Here, different sensors may contribute differently to the same perception task. Hence, it is essential to adjust the importance of different sensors to improve perception reliability and coverage. 

Building on the above discussion, we present the first wireless perception based AIGC (WP-AIGC) framework, which seamlessly integrates wireless perception with AIGC, affording users with better digital content generation services. Again, taking the virtual game as an example, the proposed framework is illustrated in Fig.~\ref{F1}. As can be seen, in the physical world, the wireless perception is used to ‘read’ user’s posture and form the skeleton image. Subsequently, the AIGC model generate the avatar and the corresponding scene according to the user's application requirements and the formed skeleton image. In WP-AIGC, the number of sensors affects the perception accuracy, i.e., the similarity between the formed skeleton and the user's actual pose, while the inference steps of AIGC affects the quality of generated content. To balance the perception accuracy and content quality, WP-AIGC receives user's feedback on satisfaction with generated content and adjusts the computing resources at edge server occupied by perception and AIGC accordingly, thereby enhancing user experience. Overall, the main contributions of this paper are as follows:
\begin{itemize}
\item We propose WP-AIGC, the first framework integrating wireless perception and AIGC for providing virtual digital services to users. The WP-AIGC includes wireless perception, AIGC based content generation, and a feedback interface, which receives user's feedback and optimizes computing resource at the edge server to enhance the user experience.

\item We propose the concept of multi-scale wireless perception, which refers to perform large-scale and small-scale perception on users in sequence. Rather than being independent, the perception of different scales assists each other by sharing the perception results, thereby improving the overall perception performance.  

\item Leveraging the collected wireless perception data and the AIGC model, we provide a practical and compelling use case to verify the feasibility of the proposed framework, lighting the way of providing virtual service via the combination of wireless perception and AIGC.
\end{itemize}

\section{AIGC and wireless perception technologies}
In this section, we provide a comprehensive review of AIGC and wireless perception technologies.
\subsection{AIGC}
The development of AIGC can be divided into three main stages. The initial stage (around 1950s to 1990s) is characterized by limited technologies, and only small-scale experiments are feasible, resulting in products such as the ‘Illiac Suite’~\cite{hiller1958musical}. In the second stage (around 1990s to 2010s), the breakthroughs in deep learning and the transformation of internet services pushed AIGC into practical applications, taking advantage of the accumulation of user data. We are now in the third stage (around 2010s to present), with the emergence of new and better AI models leading to the more powerful and intelligent AIGC, capable of producing a variety of digital content that imitates or even exceeds human creations.

\begin{figure*}[t]
	\centering
	\includegraphics[width=1\textwidth]{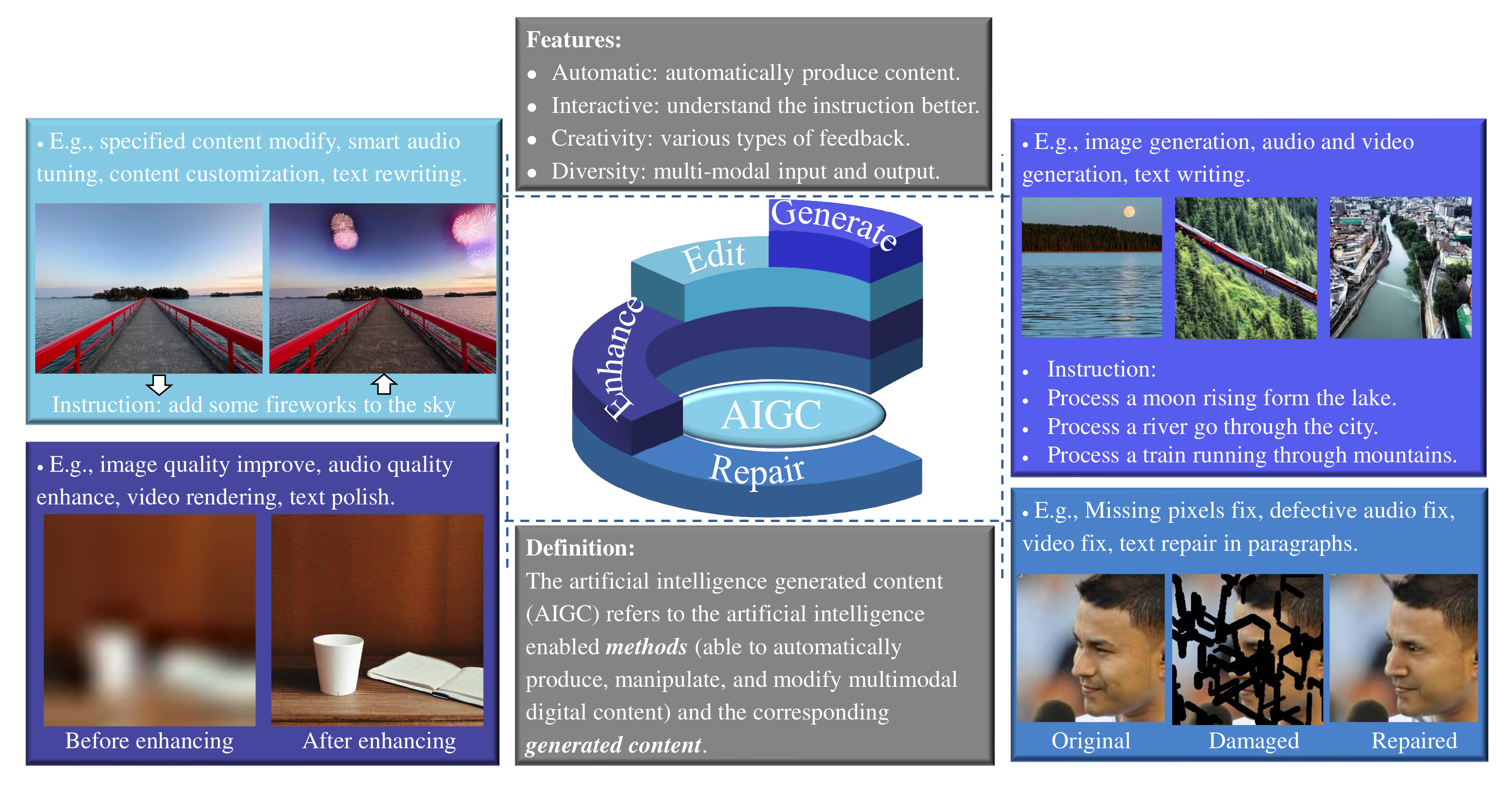}%
	\caption{The definition, features, and the corresponding applications of AIGC.}
	\label{F2}
\end{figure*}

Basic to advanced, AIGC functions includes repair, enhance, edit and generate. Specifically, repair refers to fix missing content, such as using diffusion models~\cite{dhariwal2021diffusion} to recover missing pixels in a picture. Besides that, AIGC can also enhance the image quality by increasing the contrast, improving pixel quality and clarity, as shown in Fig.~\ref{F2}. The content editing and generation are more advanced features of AIGC. Concretely, editing means that AIGC is capable of conducting operations, such modifying and replacing, on the specified content. Such a function can be used to handle the sensitive content, such as replacing sensitive people or things without changing the overall layout and style of the picture~\cite{harshvardhan2020comprehensive}. The content generation is the function of AIGC that distinguishes it from other AI models. Benefiting from a large amount of training data, algorithm progresses and hardware upgrades, the AIGC now can generate not only images, but also videos, codes, and manuscripts, which can greatly improve productivity of digital content. According to the above described definition and functions, several features of AIGC can be summarized as follows. 

\begin{itemize}
\item {\textbf{Automatic}}. Given a specified task or order, AIGC can automatically produce digital content and present it in various forms, such as pictures and videos, which is more productive than traditional ways that often requires human involvement, such as professionally generated content (PGC) and user generated content (UGC).

\item {\textbf{Interactive}}. Due to the massive training data and human-like way of thinking, AIGC can better understand the user's thoughts or instructions through input, making AIGC can interact with people in a more natural way than other AI models. Such a feature is expected to further improve the interaction experience between humans and machines.

\item {\textbf{Creativity}}. Different from the traditional AI model with limited output space in most cases (such as a classifier), AIGC have diverse answers to the same question, demonstrating its certain creative ability that can promote the diversification of digital content.

\item {\textbf{Diversity}}. The AIGC not only supports multi-modal input, but the generated digital content can also be presented in different forms, which is one of the most important characteristics that distinguish AIGC from other AI models. As a result, AIGC can assist in a variety of digital content production in different fields.
\end{itemize}
In Fig.~\ref{F2}, the definition, features and some applications of AIGC are presented. It is clear that AIGC would significantly improve the content and information production efficiency in the near future, thereby revolutionizing the traditional digital content production and consumption mode.
\subsection{Wireless Perception}
Benefiting from the ISAC technologies, using ubiquitous wireless communication signals for physical world perception gains considerable traction. The underlying principle of this approach is that the physical environment can impact the propagation of wireless communication signal, resulting in the received signal naturally containing information about the environment~\cite{du2022semantic}. By utilizing signal processing techniques, the features of the physical environment can be extracted, so as to realize the perception of the physical environment.

In the realm of wireless perception, research mainly concentrates on the detection, localization, and recognition of users as the primary target of perception. Specifically, user detection refers to the determination of the presence or absence of a user in a specific area through the analysis of fluctuations in the wireless signal. The main steps often include signal noise reduction, feature extraction and determination. The localization refers to finding the coordinates of a user relative to a reference object in physical space. This process mainly contains two key steps, including signal parameters estimation and user localization, which is typically realized by constructing geometric constraints based on the estimated parameters. The obtained user's location can be used for indoor navigation, emergency evacuation, etc. Furthermore, recognizing the user's state, behavior, gesture, and even identity is also crucial. Here, recognition is often accomplished by extracting and classifying the signal patterns. Taking user's physical activities recognition as an example, a common approach is to extract the time-frequency characteristics of signal fluctuations, and use machine learning, such as support vector machines, to achieve activity classification. We can observe that wireless perception has two notable advantages. 
\begin{itemize}
\item The first is the wide coverage. Theoretically, wireless perception can be applied at anywhere covered by wireless signals. However, wireless signals may differ across different situations, resulting in varying perception capabilities. 

\item The second is that users are not required to bear any device during the perception process, eliminating both the discomfort associated with prolonged device wear and the need to be concerned with charging. 
\end{itemize}

Based on foregoing discussion, with particular emphasis on the role of AIGC, the operating principle and advantages of wireless perception, it is natural to consider using the output of wireless sensing as the input of AIGC, thus providing users with more automatic and diverse mixed reality service. Yet, integrating these two technologies poses important technical challenges, which are discussed next.

\section{The proposed framework}
This section presents the proposed WP-AIGC framwork, including research challenges and implementation processes.
\subsection{Research Challenges}
\subsubsection{\textbf{Multi-scale wireless perception}}
Different scales of wireless perception contribute differently to the generated content. For instance, in Fig.~\ref{F2}, the user's location in the physical world determines the virtual character's position in the boxing ring, while behavior facilitates virtual character construction. Therefore, multi-scale perception is vital for WP-AIGC. This involves perceiving various types of information related to the user using the same set of data, while considering the computational resources consumed by each perception task. Furthermore, these tasks must be mutually reinforcing to enable multi-scale perception of targets with limited data.
\subsubsection{\textbf{Balance of computing resources}}
Both perception and AIGC require computing resources, and hence another key issue is to balance the resource consumption. Allocating more resources to perception can increase perception accuracy but reduce the resources available to AIGC, leading to lower quality content generation. Conversely, excessive resource consumption by AIGC may result in insufficient perception accuracy, leading to the generation of content that fails to meet users' application requirements. In practice, the requirements for generated content vary among individual users. Therefore effective resource balance depends on user feedback.
\subsubsection{\textbf{Content quality assessment}}
The evaluation of content quality is not only directly related to the QoS of the framework, but also can guide the allocation of computing resources. The content quality evaluation should consider two major aspects. The first is the accuracy of the generated content, which depends on the wireless perception accuracy and whether the user's requirements are accurately transmitted to AIGC. The second aspect is the quality of content presentation, such as whether images are clear or videos are smooth. The evaluation of the first aspect depends more on user feedback, while the second aspect can be analyzed through numerical analysis.
\subsubsection{\textbf{Mutual enhancement between perception and AIGC}}
The proposed framework involves perception and AIGC, which are independent but require mutual promotion to enhance the overall performance of the framework.  From the perspective of promoting AIGC through perception, with a certain level of perception accuracy and data refresh rate, the fewer resources occupied by perception, the more resources AIGC can use, resulting in more accurate and higher quality content generation. Similarly, improving AIGC efficiency can provide perception with more computing resources and thus enhance its performance to some extent. Besides, there are other more in-depth approaches that can be adopted, such as adjusting perception algorithms and strategies by evaluating the accuracy of the generated content.
\subsection{The Proposed WP-AIGC}
As shown in Fig.~\ref{F1}, the proposed WP-AIGC framework can be divided into three main parts. The first part is the multi-scale perception of user in physical world, which employs a series of techniques such as signal processing and machine learning to transform wireless signals into user skeleton image. The second part is the digital content generation, which creates corresponding digital content based on user needs and skeleton image. The third part is the feedback control, which adjusts the allocation of computing resources at edge server according to user feedback, thereby achieving a balance in performance.

\subsubsection{Multi-scale sensing}
\begin{figure*}[t]
	\centering
	\includegraphics[width=1\textwidth]{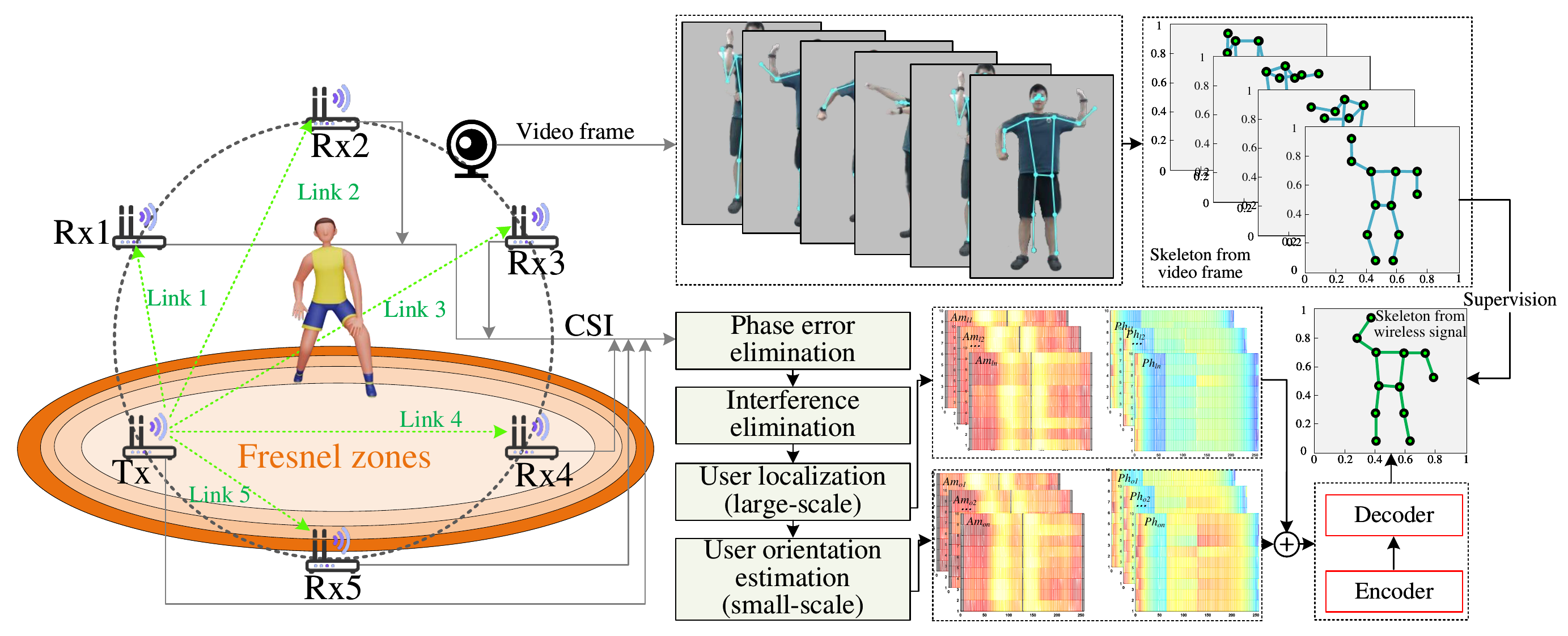}%
	\caption{The multi-scale perception and training process of the network that maps signals to skeleton image. Here, TX is the wireless signal transmitter, and RX is the receiver. The green dashed lines are the transmission links and the orange concentric circles represent the Fresnel zone formed by TX and RX4. During training, wireless signals and video signals are collected synchronously. The CSI amplitude and phase vectors are then extracted from large to small scales and fed to the network for predicting the skeleton. Simultaneously, the human skeleton extracted from the video frames is used as the ground truth to optimize the skeleton predicted by the wireless signals, so as to completing the training. During the operation, the trained network can directly convert CSI into skeleton images without the assistance of a camera.}
	\label{F3}
\end{figure*}

A straightforward method for multi-scale perception is to use different algorithms to process the data, to realize various perception purposes. Yet, during this process, the perception algorithms operate independently and as such are unable to promote each another. To improve perception accuracy, we propose to perceive users in order of large-scale to small-scale, and enhance the perception accuracy by considering the results of large-scale perception in the process of small-scale perception, as shown in Fig.~\ref{F3}.
Specifically, assuming there are several multi-antenna wireless devices with known locations participating in the perception, where one of them plays as transmitter and the rest are receiver. Based on the received signal, the channel state information (CSI) is extracted and the phase error is eliminated via conjugate multiplication operation. 

As the direct signal and static objects introduced reflections do not contain information about the user, and hence the elimination of such signals is performed after phase error correction. After that, we use the matrix pencil algorithm to estimate the angle of arrival (AoA) and time of flight (ToF) of the user induced reflection, and then utilize the estimation results to build the constraints and determine the user's location (large-scale information). On this basis, we weight the CSI of links based on the distance between the user and the link (closer links receive higher weights) and sum the amplitude and phase of the CSI for all links, to get the amplitude and phase vectors, respectively.

The orientation of the user relative to the link affects the accuracy of small-scale perception. Based on the user's location, therefore, further analysis is conducted on the Fresnel zone formed by the surrounding links of the user. The Fresnel zone~\cite{wang2016human} consists of a series of concentric ellipses. When an object moves perpendicular to the elliptical boundary, it passes through more ellipses, causing greater fluctuations in the received signal. Conversely, when it moves parallel to the boundary, the signal fluctuations are smaller. Thus, by analyzing the signal fluctuations of different links, the user's orientation can be determined. Leveraging the obtained orientation, the amplitude and phase of the CSI of the links are weighted and summed, to obtain another pair of vector for for user pose perception (small-scale).

\subsubsection{Signal to skeleton conversion}
Using the obtained vectors, a neural network is trained to map the extracted amplitude and phase vectors to human skeletal images. To this end, we simultaneously use the camera to capture the video stream when collecting CSI training data, and extract the human skeleton from the video frame as the ground truth for the training process. Considering temporal and spatial correlations of user's state, we aggregate multiple amplitude and phase vectors as inputs and use the convolutional neural networks as basic blocks to construct the encoder and decoder to realize mapping process.

Concretely, The encoder network consists of three layers of 3x3 convolutions with 2x2 strides, each followed by a 1x1 convolutional layer with a stride of 1x1. The ReLu activation functions are applied after each layer and a fully connected layer is utilized after the final convolutional layer to directly convert images. The decoder network consists of a total of seven layers, where the first two layers use 1x1 kernels with a stride of 1x1, and the subsequent 5 layers use 3x3 convolutions with a stride of 1x1. The network is implemented using TensorFlow and trained over 64 epochs with a batch size of 32, with a learning rate of 0.001. Notably, the encoder network and decoder network were jointly trained. During the training phase, the network takes aggregated amplitude and phase vectors as inputs and produces a corresponding predicted human skeletal image. Then, the neural network is optimized under the supervision of skeleton extracted from video frames. The training objective is to minimize the difference between the predicted skeleton and the corresponding ground truth skeleton, with the loss function defined as the average euclidean distance error.

\subsubsection{Digital content generation}
After converting wireless signals into skeleton image, the InstructPix2Pix~\cite{brooks2022instructpix2pix},  which can edit images based on human instructions, is utilized to generate corresponding virtual digital content based on the user's requests and the obtained skeletal images. 

The construction of InstructPix2Pix mainly consists of two steps: generating training data and training the model. During the first step, a fine-tuned GPT-3 is used to generate instructions and edited captions, and then StableDiffusion~\cite{rombach2022high} is combined with PrompttoPrompt~\cite{hertz2022prompt} to generate paired images based on the paired instructions and captions. In the second step, based on the generated pairs of images and corresponding instructions, a network is trained to predict the noise added to the noisy latent given the image conditioning and the text instruction. During the training process, the available weights of the InstructPix2Pix model are initialized with a pre-trained stable diffusion checkpoint. Meanwhile, to enable the trained model to perform conditional or unconditional denoising with respect to either or both conditional inputs, 5\% of the sample images, 5\% of the sample instructions, and 5\% of the sample images with instructions are randomly set to an empty set. Based on such capability, two scale guidance parameters are further introduced to trade off how strongly the generated samples correspond with the input image and how strongly they correspond with the edit instruction. After completion of the training, WP-AIGC utilizes this network to generate digital content, with the skeleton image from wireless perception as input and user's service requirements as editing instructions to meet desired specifications.

\subsubsection{Computing resource adjustment} Besides the above mentioned modules, the WP-AIGC framework features an interface for collecting user feedback, which is used to adjust computational resources and improve QoS. During the initial stages of operation, the WP-AIGC allocates computational resources according to a default ratio that considers factors such as the number of devices involved in perception and the minimum number of inference steps required. Meanwhile, Total Variation (TV) and Blind/Referenceless Image Spatial Quality Evaluator (BRISQUE) are used to monitor the quality of the generated images and ensure that they meet the basic criteria~\cite{du2023enabling}. Here, the TV characterizes the smoothness of the images, while BRISQUE quantifies the potential loss of ``naturalness" in the images. On this basis, WP-AIGC optimizes the available resources once it receives feedback from users. For example, if the feedback reveals a disparity between the posture of the generated virtual character and user's actual posture, then WP-AIGC increases the resources allocated to perception to enhance perception accuracy.

\section{Case study}
To validate WP-AIGC, we conducted tests in a practical scenario. Specifically, we built a wireless perception platform based on the 801.11ac protocol. This platform consists of a signal transmitter and five receivers for signal reception and CSI extraction. The receivers forwarded the extracted CSI to the edge server, which used multi-scale perception and a pre-trained neural network to convert the CSI into the user's skeletal images in a non-real-time manner. On this basis, the skeleton image and user's service requests are fed to AIGC as guidance to generate corresponding content.

\begin{figure}
    \centering
    \includegraphics[width=3in]{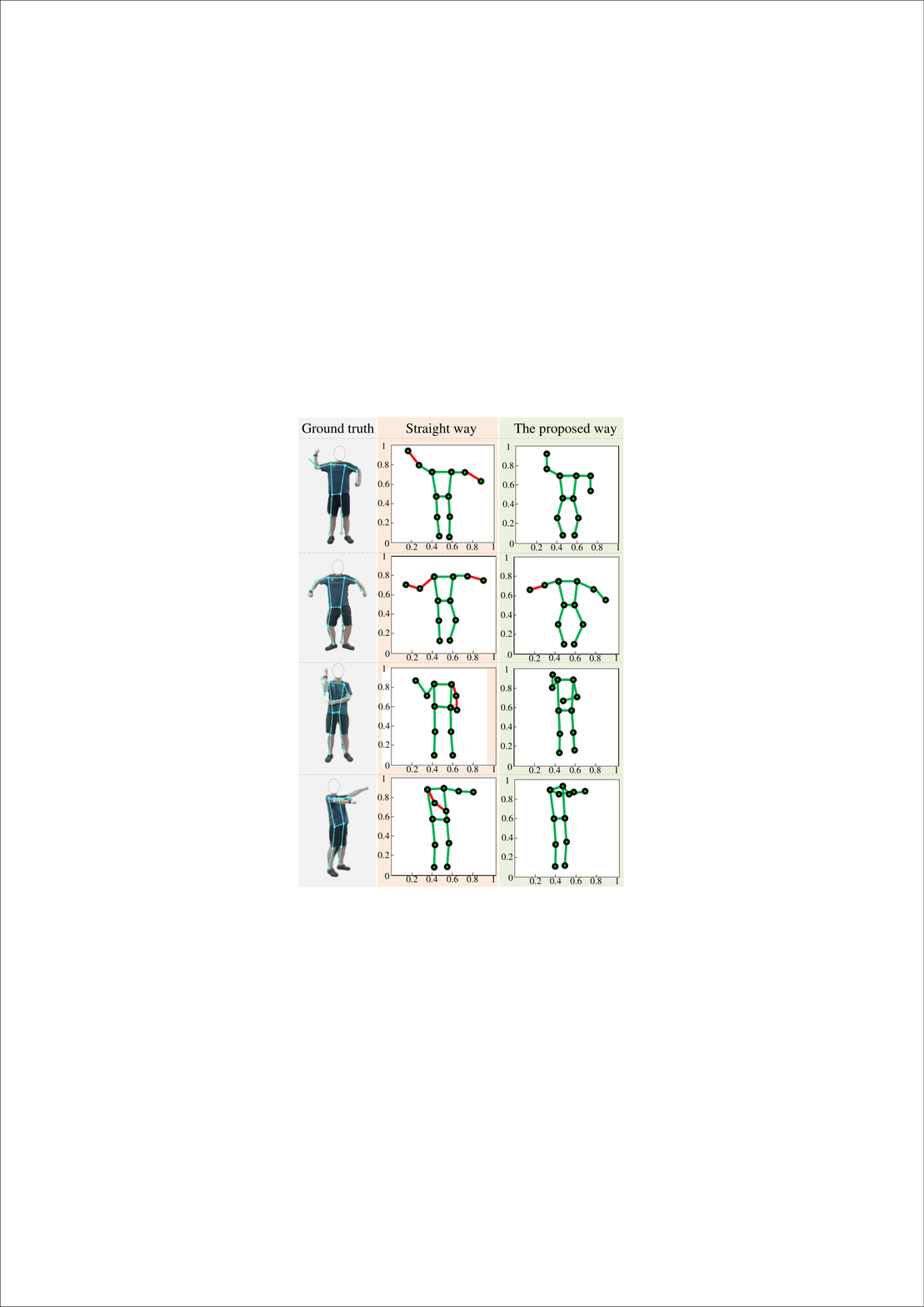}
    \caption{Comparison of experimental results of different perception methods.}
    \label{F4}
    \end{figure}

To begin with, we validate the advancement of the proposed multi-scale perception by comparing the skeleton images generated with and without the use of multi-scale perception technology. As illustrated in Fig~\ref{F4}, the overall similarity between the skeletons generated by the two methods and the real human pose is relatively close, but there are still certain differences in details, such as the red parts in the skeletal images. Particularly, in the case that the arms are close to the torso, such as the pose corresponding to the third row of images, it is difficult to distinguish the signals reflected from the torso and those from the arms. In this situation, multi-scale perception produces better results.

\begin{figure*}[t]
	\centering
	\includegraphics[width=1\textwidth]{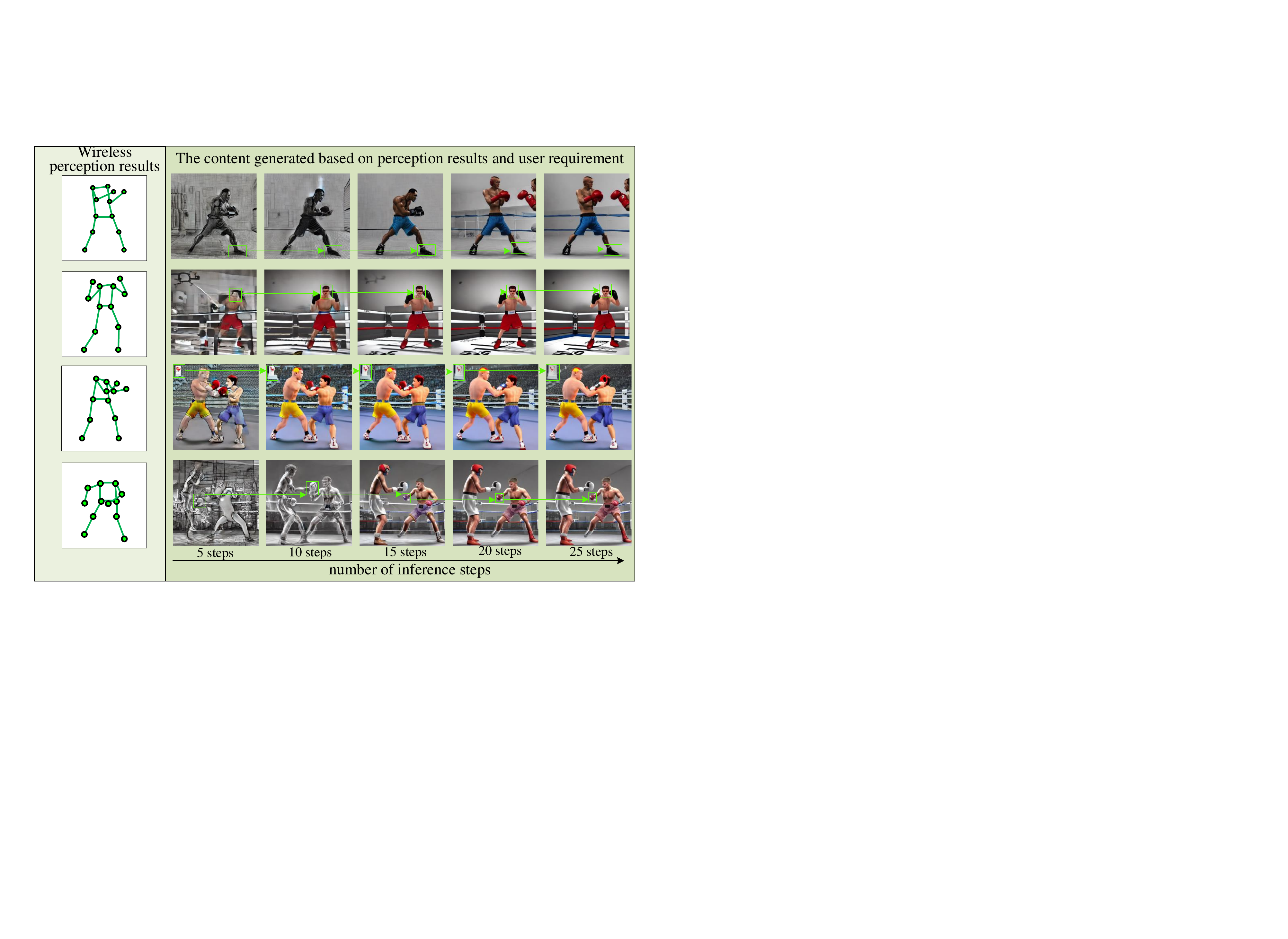}%
	\caption{The influence of AIGC steps on the generated content. The image on the left are the input skeleton image, while the image on the right represents the corresponding generated content. As can be seen, more steps trigger more natural content, as denoted by the green boxes.}
	\label{F5}
\end{figure*}

Following the validation of multi-scale perception, we conduct an analysis on image generation, using boxing as an example. The results are presented in Fig~\ref{F5}, which reveals that, firstly, the model is capable of generating images based on skeleton image and user's instruction, affirming the effectiveness of the entire WP-AIGC architecture. Secondly, under a constant skeleton data and instruction set, an increase in the number of inference steps leads to a significant improvement in the image quality. Such an improvement manifests itself in two key aspects: enhanced clarity, as indicated by the virtual character's facial features; and augmented realism, as demonstrated by the more natural color scheme of the boxing gloves corresponding to the character in the image located at the fourth row and third column. Hence, the abundance of computing resources allocated to AIGC is expected to correspondingly improve the generated content quality.

\begin{figure}
    \centering
    \includegraphics[width=3.5in]{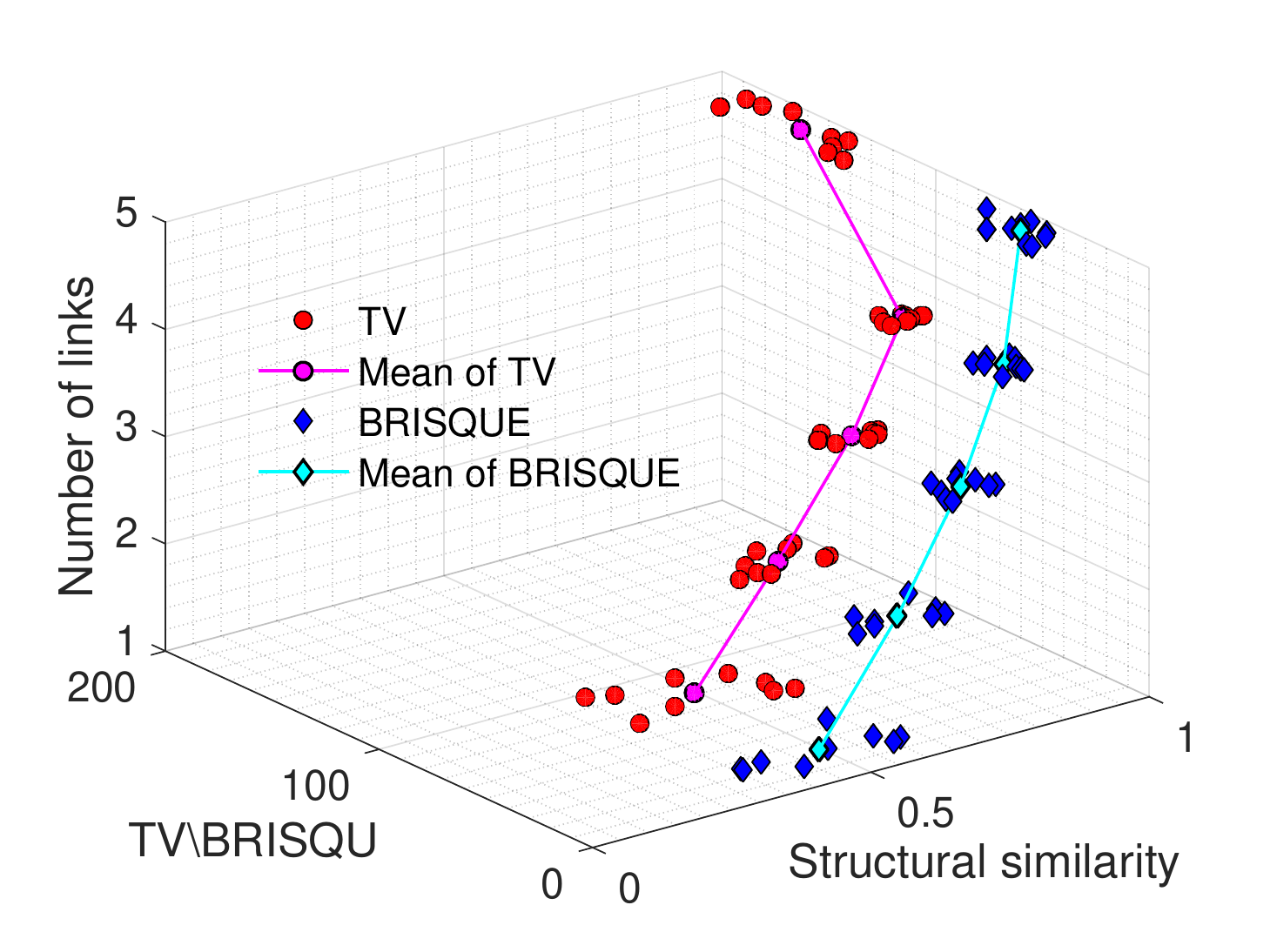}
    \caption{Effects of link quantity on structural similarity, TV, and BRISQUE.}
    \label{F6}
    \end{figure}

However, in practice, the overall computing resources are often limited, necessitating a balance between the resources allocated to perception and AIGC. Thereby, we explore the relationship between perception accuracy (evaluated by the structure similarity between the generated skeleton and the real pose) and image quality (measured using TV and BRISQUE) under the constraint of content refresh rate at 1 Hz. Specifically, assuming a processing time of 0.1 seconds for one perception link and 0.05 seconds for generating skeleton images, then the time for AIGC inference in one time slot is given by $\left( {1{\text{ - }}\# \left( {links} \right) \times 0.1 - 0.05} \right)s$, where $\# \left( {links} \right)$ is the the number of links.  Based on above assumptions, the relationship between perception accuracy and image quality is illustrated in Fig.~\ref{F6} 

The experimental results show that increasing the perceptual link improves the perception accuracy, i.e., the similarity is increased. However, it reduces the computing resources of AIGC, resulting in an increase in TV and BRISQUE, that is, a decrease in image quality. In contrast, reducing the perceptual links lowers the perceptual accuracy, but improves the image quality. Given that user's demands for perceptual accuracy and image quality may vary across different applications, such results evidently show that it is crucial to obtain user feedback and adjust computational resources in situations where resources are limited. Meanwhile, the result provides a reference for the initial allocation of resources.

\section{Future Directions}
\subsection{The AIGC Model Selection Strategy}
Nowadays, various models are available for generating virtual digital content, each of which excels in different domains and consumes different amounts of computing and communication. Therefore, it is crucial to design an AIGC model selection mechanism that carefully considers user requirements, resource availability, and other relevant factors. Furthermore, in situations where historical data is available, one can leverage user feedback to gain insights into user preferences and customize the AIGC to the user. For example, if a user likes to switch virtual characters while playing virtual game, then we should choose an AIGC model that excels in generating virtual characters from the pool of models capable of generating virtual game content to further enhance QoS.

\subsection{Edge Computing Resource Optimization}
In deploying AIGC in mobile networks, the edge computing resource optimization needs to be considered. The resources here include computing resources required for digital content generation, storage space occupied, and transmission resources consumed for content delivery, such as bandwidth and transmitting power~\cite{du2023enabling}. The joint resource optimization model design process must ensure content quality and minimize response time delays, while also preventing overload of the AIGC model due to excessive tasks. Additionally, the impact on other network services must be taken into account. One potential approach is to utilize deep reinforcement learning to allocate various network resources.

\subsection{Guiding AIGC with Other Promising Techniques}
Regarding the proposed WP-AIGC, further research can be conducted on how to expand the range that users can move around and the number of users, so as to improve the system's practicality. Besides, AIGC can be combined with other promising technologies, such as eye-tracking and brain-computer interfaces~\cite{wu2020transfer}, to guide or even control AIGC in generating more complex digital content that better meets user's requirements. However, signals such as human brain waves are weaker, and are also influenced by human emotions and thoughts, making them contain more information. Therefore, when using brain waves to guide the AIGC, the difficulty lies in accurately inferring user needs based on raw signals and efficiently transmitting these needs to the AIGC model. Meanwhile, ensuring the security of both the original signal and the generated content is crucial, and blockchain can be introduced here to achieve these goals.

\section{Conclusions}
In this article, we have proposed the WP-AIGC framework which leverages the multi-scale wireless perception technology to sense users in the physical world. It then sends the perception results to the AIGC model, which generates digital content according to the user’s service requests. The effectiveness of the proposed framework has also been verified through experiments. Unlike most existing works that take descriptive text as input, WP-AIGC takes perception results as input, which can more accurately convey the user's behavioral and status information in physical space to AIGC, resulting in a tighter connection between the generated content and the user. Additionally, in designing WP-AIGC, we also considered the issue of balancing computing resources. To our knowledge, WP-AIGC is the first framework to combine wireless perception with AIGC technology to achieve digital content generation. For the future work, we will further investigate specific technical issues in the process of combining perception with AIGC, thereby further unleashing the productivity of AIGC.
\bibliographystyle{IEEEtran}
\bibliography{Ref.bib} 

\end{document}